\documentclass[10pt, a4paper]{article}
\usepackage{tabularx}
\usepackage{makecell}
\usepackage{booktabs}
\usepackage{multirow}
\usepackage{ragged2e}
\usepackage{caption}
\usepackage{amsmath} 
\usepackage{xcolor}

\usepackage{lrec2026} 

\title{EncouRAGe: Evaluating RAG Local, Fast, and Reliable}

\name{Jan Strich, Adeline Scharfenberg, Chris Biemann, Martin Semmann} 

\address{Universität Hamburg}

\abstract{
We introduce \textbf{EncouRAGe}, a comprehensive Python framework designed to streamline the development and evaluation of Retrieval-Augmented Generation (RAG) systems using Large Language Models (LLMs) and Embedding Models. EncouRAGe comprises five modular and extensible components: \textit{Type Manifest}, \textit{RAG Factory}, \textit{Inference}, \textit{Vector Store}, and \textit{Metrics}, facilitating flexible experimentation and extensible development. The framework emphasizes \textbf{scientific reproducibility}, \textbf{diverse evaluation metrics}, and \textbf{local deployment}, enabling researchers to efficiently assess datasets within RAG workflows.
This paper presents implementation details and an extensive evaluation across multiple benchmark datasets, including \textit{25k QA pairs} and \textit{over 51k documents}. Our results show that RAG still underperforms compared to the \textit{Oracle Context}, while \textit{Hybrid BM25} consistently achieves the best results across all four datasets. We further examine the effects of reranking, observing only marginal performance improvements accompanied by higher response latency.
\\
\Keywords{Retrieval-Augmented Generation, Evaluation, Library}
}

\begin{document}
\hyphenation{En-cou-RAGe}
\maketitleabstract

\section{Introduction}

Retrieval-Augmented Generation (RAG)~\citep{lewis_retrieval-augmented_2020, huang_survey_2024, nikishina_creating_2025} has emerged as a dominant approach for enhancing large language models (LLMs) with external knowledge sources. By combining retrieval and generation, RAG systems can provide more factual, contextually grounded, and up-to-date responses \citep{shuster_retrieval_2021, sahoo_comprehensive_2024}. 
However, the rapid development of RAG architectures \citep{huang_survey_2024, sarthi_raptor_2024}, methods \citep{ gao_complementing_2021, gao_precise_2023}, and LLM-as-a-Judge metrics \citep{es_ragas_2024} has created a need for an implementation-focused RAG evaluation library to create a reliable way to compare performance on domain-specific datasets.

Assessing a RAG system for a specific dataset requires analyzing the performance of both the retriever and generator, as well as their interaction. 
It is particularly challenging because LLM outputs can be lengthy, complex, and ambiguous, making it difficult to define clear correctness or objectively measure factual accuracy \citep{ru_ragchecker_2024}. 
Recent RAG evaluation frameworks, such as RAGAS \citep{saad-falcon_ares_2024}, RAGChecker \citep{ru_ragchecker_2024}, and TruLens \citep{snowflake_inc_trulens_2024}, have begun addressing these challenges by providing libraries to jointly analyze retrieval and generation quality, reflecting the interdependence between the two components. 
Instead of relying solely on traditional metrics like BLEU~\citep{papineni_bleu_2002} or ROUGE~\citep{lin_rouge_2004}, they use LLM-based or embedding-based measures to directly evaluate factual correctness, relevance, and consistency of the generated text \citep{es_ragas_2024}. Additionally, they often provide enhanced UI and monitoring features for RAG in production environments.

Despite recent advances, a lack of the following features in state-of-the-art (SOTA) frameworks remains, hindering the effective comparison of RAG methods for domain-specific datasets: \\
\textbf{(1) Lack of comparability of scientific methods.}
Existing libraries in this field focus solely on implementing metrics, with no standardized implementation of SOTA RAG methods. This heterogeneity hinders meaningful and reproducible comparisons across different RAG approaches. \\
\textbf{(2) Limited flexibility of LLM-as-a-judge metrics.}
LLM-as-a-judge metrics are sensitive to domain and dataset, and must be flexible to be effective. Therefore, custom prompts for individually designing evaluations are crucial and are not currently available in existing frameworks.
\\
\textbf{(3) Primarily cloud-oriented, with limited flexibility for integrating additional metrics or methods.}
Many existing frameworks are designed with a narrow focus on specific local setups and lack modularity for broader integration. A flexible, plug-and-play architecture is essential to support new RAG methods and additional new metrics. 

\begin{figure*}[!ht]
\begin{center}
\includegraphics[width=1\textwidth]{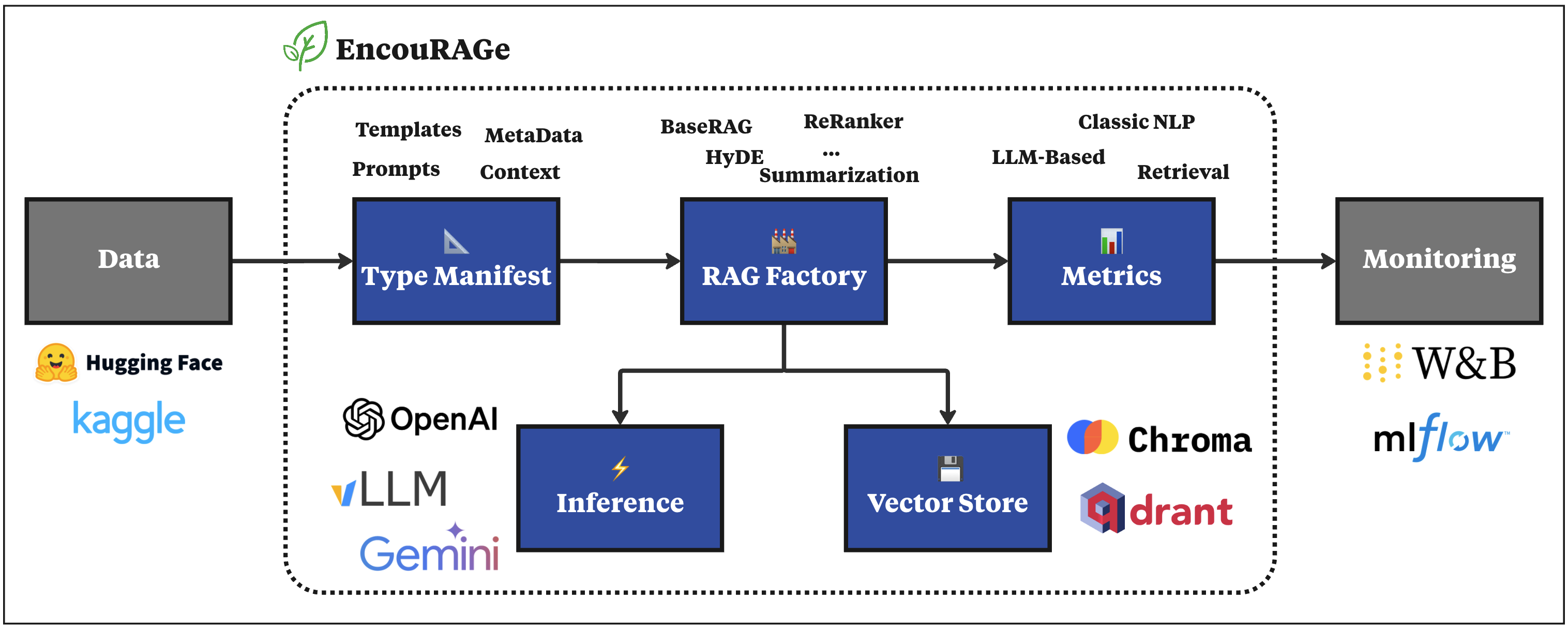}
\caption{System overview of the \textbf{EncouRAGe} Python library. Input data in any format must be transformed to fit the \textbf{Type manifest}. The \textbf{RAG Factory} provides various RAG methods, while the \textbf{Metrics} component implements all evaluation metrics. \textbf{Inference} can be executed locally or via cloud providers using the OpenAI SDK, and supported \textbf{Vector stores} include Chroma and Qdrant. The output fits popular monitoring systems.}
\label{fig:overview}
\end{center}
\end{figure*}

To tackle these challenges, we present \textbf{EncouRAGe}, an open-source Python library for comprehensive, reproducible, and extensible RAG evaluation (Figure~\ref{fig:overview}).
It integrates any dataset into a parsable, object-oriented structure for effective and traceable workflows.
EncouRAGe includes 10 RAG methods, manages LLM and embedding inference and provides metrics in three categories:
\begin{enumerate}
\item \textbf{Generator Metrics}: Evaluating the performance end-to-end with classic NLP metrics (ROUGE, BLEU, etc.).
\item \textbf{Retrieval Metrics}: Evaluating the effectiveness of the retriever in finding relevant information from the knowledge base.
\item \textbf{LLM-as-a-Judge}: Evaluating the effectiveness of the whole RAG Pipeline enhanced by custom LLM prompts.
\end{enumerate}

To address the rapid pace and volume of new RAG and LLM research, reproducing and comparing methods has become increasingly time-consuming.
With EncouRAGe, evaluating new approaches or verifying others claims becomes straightforward and extensible.
The framework enables quick benchmarking across diverse methods and datasets with minimal manual effort, requiring mainly GPU time rather than complex reimplementation.
We demonstrate this through experiments comparing multiple RAG methods on four popular QA datasets.
The datasets and code are open source and anonymized for submission.\footnotemark
\\
The main contributions of this paper are as follows:
\begin{itemize}
    \item We present \textbf{EncouRAGe}, a novel RAG evaluation framework that consists of \textbf{10} SOTA \textbf{RAG methods}, an object-oriented \textbf{type manifest}, and over \textbf{20 metrics} to get transparent and reliable results.
    \item We conduct a \textbf{comparison of RAG methods} using EncouRAGe on four datasets with \textbf{25k QA pairs} and \textbf{50k documents}, demonstrating the effectiveness and usefulness of the framework.
    \item We further present one ablation study, showing the effectiveness of \textbf{Reranker} ratio for two cross-encoder models and a subset of the four datasets.
    
\end{itemize}

\footnotetext{\url{https://anonymous.4open.science/r/encourage-B501/}}

\section{Related Work} \label{sec:related_work}

\subsection{Retrieval-Augmented Generation} \label{subsec:rag}

LLMs excel at text generation but face challenges such as outdated knowledge and hallucinations~\citep{shuster_retrieval_2021, huang_survey_2024}. RAG addresses these issues by incorporating external knowledge, improving accuracy and factuality \citep{lewis_retrieval-augmented_2020, sarthi_raptor_2024}. This approach is critical in domains such as law~\cite{cui_chatlaw_2024}, medicine~\cite{xiong_benchmarking_2024, chen_towards_2025}, and finance~\cite{chen_towards_2025}, where precision is crucial. RAG systems have achieved strong results in tasks such as open-domain question answering~\cite{kim_re-rag_2024}, code generation~\cite{wang_code_2025}, and dialogue~\cite{chen_wavrag_2025}. They have also been successfully adopted in real-world applications, including LlamaIndex~\cite{liu_llamaindex_2022}.

\subsection{Evaluation of RAG} \label{subsec:evaluation_rag}

Evaluation RAG systems can be categorized into three domains: tools for lightweight and dataset-based evaluation, monitoring frameworks for production applications, and general-purpose evaluation frameworks for LLMs.

\textbf{(1) Lightweight and Dataset-Based Evaluation Tools.}
RAGAS~\citep{es_ragas_2024} and RAGChecker~\citep{ru_ragchecker_2024} provide systematic metrics for assessing RAG pipelines. RAGAS was the first library to provide LLM-as-a-Judge metrics and offers additional classic NLP evaluation metrics. In contrast, RAGChecker employs a modular approach that decomposes RAG performance into retrieval and generation components, allowing detailed error attribution and interpretability. But both lack RAG method implementation and are heavily cloud-focused.

\textbf{(2) Monitoring Frameworks for Production RAG Applications.}
TruLens~\citep{snowflake_inc_trulens_2024} and Opik~\citep{comet_ml_inc_comet_2025} extend evaluation capabilities to deployed RAG and LLM-based applications. TruLens enables real-time tracking, custom evaluation functions, and integration with human feedback loops, making it particularly suited for agent-oriented systems where RAG acts as an auxiliary mechanism. Opik provides end-to-end monitoring and analytics for production environments, supporting model observability for improvement workflows. Both systems are application-focused and not designed for research.

\textbf{(3) General LLM Evaluation Frameworks.}
DeepEval~\citep{confident_ai_deepeval_2024} and OpenAI evals~\citep{openai_openai_2025} offer broader evaluation capabilities that extend beyond RAG-specific scenarios. DeepEval provides standardized testing interfaces and model-agnostic benchmarking tools for assessing reasoning, factual accuracy, and robustness. OpenAI’s evals framework enables reproducible and scalable evaluation of LLMs through configurable metrics and task templates.

\section{EncouRAGe} \label{sec:encourage_framework}
As shown in Fig.~\ref{fig:overview}, EncouRAGe is an end-to-end RAG evaluation library in Python consisting of five main components. To sharpen the scope of our framework, we define a task in Subsec.~\ref{sec:task} and provide an overview of each component to illustrate the framework’s idea and workflow.
EncouRAGe supports any dataset containing queries, answers, and golden context triples. The data is standardized through the \textit{Type Manifest} as explained in Subsec.~\ref{subsec:type_manifest}, ensuring type safety and correct assignment of all elements during processing.
Users can select from a wide range of RAG methods defined in the \textit{RAG Factory} (Subsec.~\ref{subsec:rag_factory}), which utilize the \textit{Inference Handler} and \textit{Vector Store } described in Subsec.~\ref{subsec:inference_vector_store}.
For the main evaluation, EncouRAGe uses over 20 \textit{metrics} across three categories to transparently understand how well RAG methods perform and to identify whether performance loss originates from the retriever or the generator.
EncouRAGe features a fully tested codebase and supports all LLMs and embedding models available on Hugging Face and all models implementing the OpenAI SDK.

\subsection{Task Formulation}\label{sec:task}
To formally describe the idea and foundation of our framework, which serves as the basis of our library, we define the following mathematical formulations. 
\textbf{EncouRAGe} provides a collection of RAG methods \(M\) and evaluation metrics \(\mathcal{M}\), which can be applied to any dataset that satisfies the assumptions of \(\mathcal{D}\).
Let a dataset be defined as 
\[
\mathcal{D}= \{(q_i, a_i, c_i)\}_{i=1}^{N},
\]
where \(q_i\) denotes a question, \(a_i\) its corresponding answer, and \(c_i\) the associated gold context.
All contexts \(c_i\) are embedded and stored in a vector store
\[
\mathcal{V} = \text{Embed}(\{c_i\}_{i=1}^{N}),
\]
which serves as the retrieval space for RAG methods.
A RAG method is denoted as 
\[
M = (R, G),
\]
where \(R\) represents the retriever and \(G\) the generator.
Given a query \(q_i\), the retriever \(R\) accesses the vector store \(\mathcal{V}\) using the retrieval strategy defined by \(M\) to return a set of top-k relevant contexts
\[
C_i = R_k(q_i; \mathcal{V}, M),
\]
where \(R_k(\cdot)\) denotes the retrieval of the \(k\) most relevant contexts from \(\mathcal{V}\) according to the similarity function specified by \(M\).
$C_i$ is then passed to the generator \(G\) to produce a predicted answer
\[
\hat{a}_i = G(q_i, C_i).
\]

The quality of \(\hat{a}_i\) is evaluated against the ground-truth answer \(a_i\) using a set of metrics 
\[
\mathcal{M} = \{m_1, m_2, \dots, m_K\},
\]
yielding performance on multiple metrics
\[
s_{i,k} = m_k(\hat{a}_i, a_i), \quad \forall i \in [1, N],\; k \in [1, K].
\]
And each metric $s_{i,k}$ helps to understand the performance of the retriever, generator, and their interaction what is the main goal of EncouRAGe.

\subsection{Type Manifest} \label{subsec:type_manifest}
\begin{figure}[!t]
\begin{center}
\includegraphics[width=1\columnwidth]{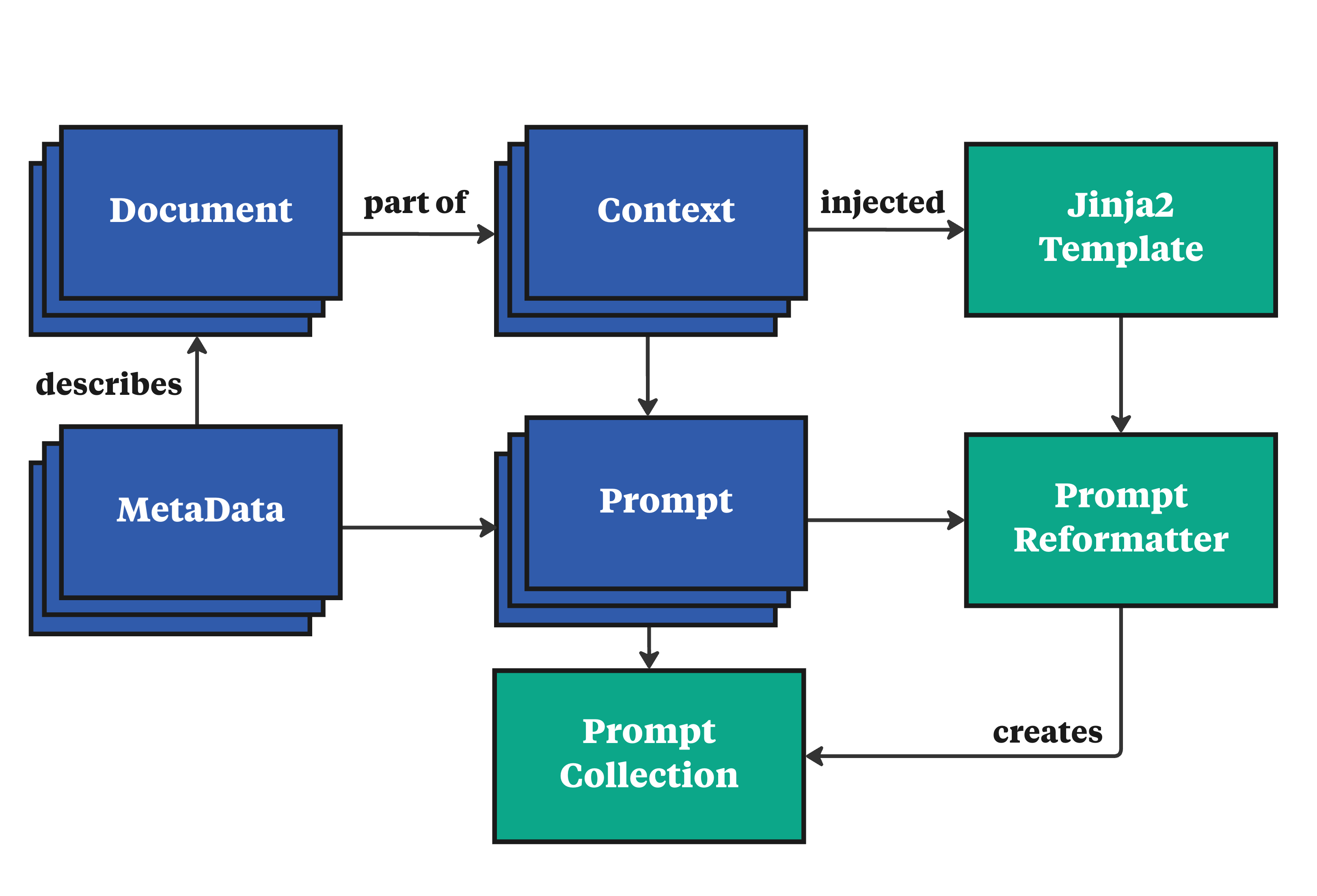}

\caption{\textit{Type Manifest} of EncouRAGe. Gold \textbf{Documents} link to the \textbf{Context}, combined via \textbf{Jinja2 template} with the prompt to form the final \textbf{Prompt Collection.} \textbf{Metadata} ensures traceability for documents and prompts.}
\label{fig:type_manifest}
\end{center}
\end{figure}

The \textit{Type Manifest} is a collection of object-oriented Python structures designed to streamline evaluation, improve reliability, and enhance maintainability, as illustrated in Fig.~\ref{fig:type_manifest}.
Each data point is represented as a \texttt{Document} object containing a unique identifier, textual content, and an associated \texttt{MetaData} object. The \texttt{MetaData} object is a dictionary for auxiliary information not directly used in processing, enabling flexible organization and key-value-based analysis afterward.
A \texttt{Context} may include multiple \texttt{Documents}, reflecting the capability of RAG methods to retrieve several documents from a vector store. Each \texttt{Context} is injected into a \texttt{Jinja2} template, allowing users to define prompt structures and dynamically incorporate document content via template variables from \texttt{Jinja2}.
The \texttt{Prompt} object contains an identifier, conversation history (including system and user messages), a \texttt{Context}, and a \texttt{MetaData} object. When constructing a \texttt{PromptCollection}, the \texttt{PromptReformatter} renders the \texttt{Jinja2} template with the specified system and user prompt, and the related injected template variables. This unified prompt collection ensures consistent usage across the evaluation pipeline, allowing seamless integration with the \textit{RAG Factory} and stable, reproducible metric computation.

\vspace{-0.3cm}
\subsection{RAG Factory} \label{subsec:rag_factory}

\begin{figure}[!t]
\begin{center}
\includegraphics[width=1\columnwidth]{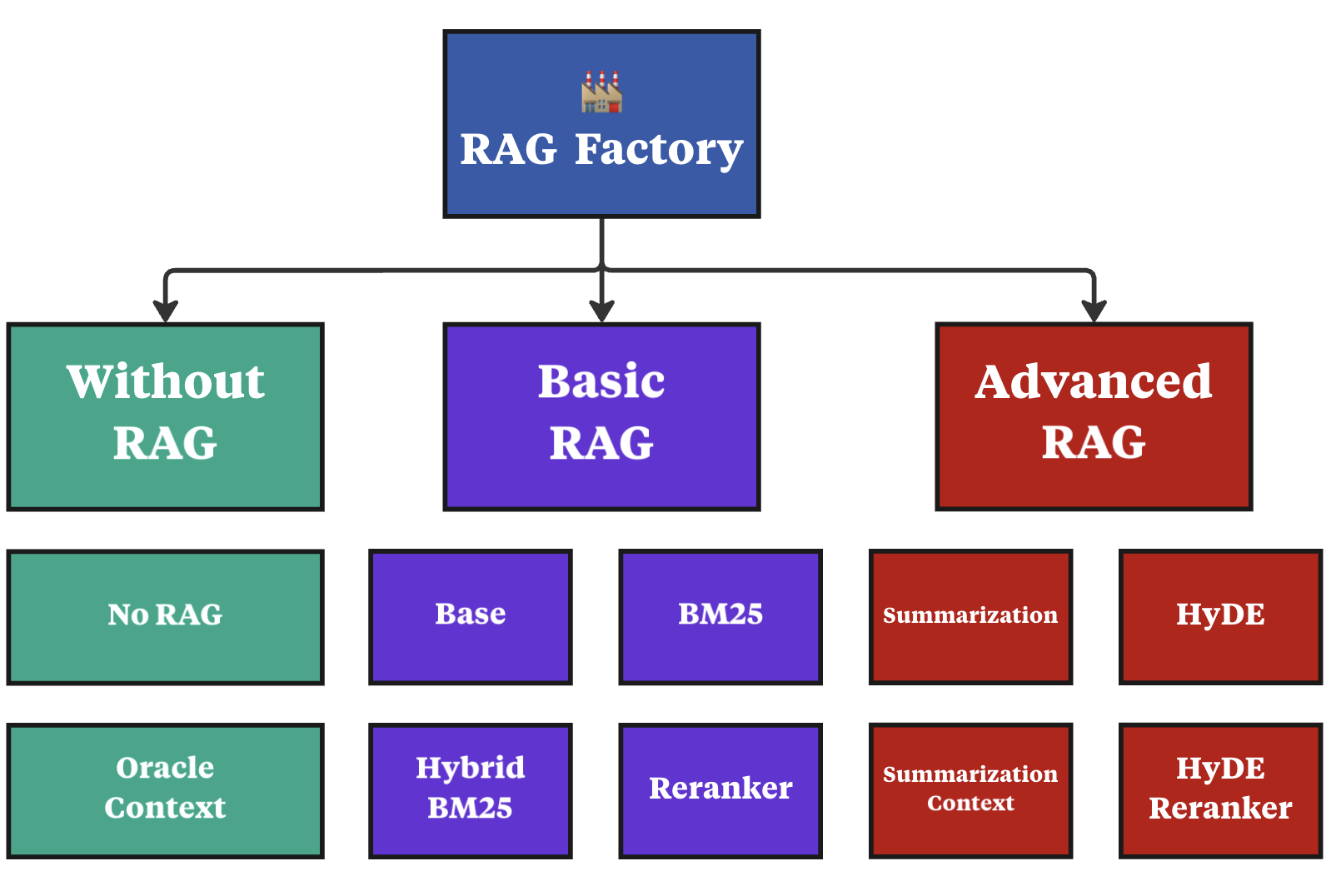}
\caption{Overview of \textit{RAG Factory} in EncouRAGe. RAG Factory is organized into three categories: Without RAG, Basic RAG, and Advanced. In total, EncouRAGe supports  \textbf{10 methods}, with more to be added in the future.}
\label{fig:rag_factory}
\end{center}
\end{figure}

The \textit{RAG Factory} constitutes the core of our Python library, defining all available methods for dataset evaluation as shown in Fig.~\ref{fig:rag_factory}. Currently, it comprises ten methods grouped into three categories: \textbf{Without RAG}, \textbf{Basic RAG}, and \textbf{Advanced RAG}. The Without RAG category includes methods that assess performance either without any contextual information or with access to golden documents, thereby illustrating the performance gap between having no knowledge and complete knowledge. This comparison indicates the potential performance improvement RAG methods can achieve when the correct context is not known in advance. The distinction between Basic and Advanced RAG lies in the additional step in Advanced RAG, which uses an LLM to generate intermediate outputs. Overall, the library offers a standardized Python interface that facilitates the addition of new methods, while the \textit{Type Manifest} ensures their smooth and consistent integration.

\paragraph{Without RAG.}
In the \emph{Pretrained-Only} setup, no retriever is employed, and models must answer questions solely based on their pretraining knowledge. Conversely, the \emph{Oracle Context} setting assumes that the relevant context is directly passed to the generator.

\paragraph{Basic RAG Methods.}
This category includes approaches that retrieve documents using standard embedding-based methods. The \emph{Base RAG} implementation follows the original RAG approach~\cite{lewis_retrieval-augmented_2020}, where only the question is embedded to retrieve the top-k documents, which are then passed unchanged to the generator. In addition, a \emph{Standard BM25}~\cite{robertson_probabilistic_2009} retriever is provided as a purely lexical baseline, relying on term-frequency and inverse document-frequency weighting to rank documents based on keyword overlap. \emph{Hybrid BM25}~\cite{gao_complementing_2021} combines sparse lexical retrieval using BM25 with dense vector retrieval, leveraging both methods to improve recall and relevance. Finally, the \emph{Reranker} method~\cite{tito_document_2021} can apply any cross-encoder model after initial retrieval to reorder documents based on their significance in a shared embedding space.

\paragraph{Advanced RAG Methods.}
This category consists of methods that modify the query, transform retrieved contexts, or employ iterative retrieval strategies. The \emph{HyDE} method~\cite{gao_precise_2023} generates hypothetical answers for each question, using them as refined queries to retrieve more relevant documents. \emph{Summarization} reduces noise by summarizing each retrieved context using an LLM, focusing on essential information. \emph{SumContext} applies a similar summarization step, retaining the original full documents for generation, aiming to reduce distractions while preserving content fidelity.

\subsection{Inference and Vector Store} \label{subsec:inference_vector_store}
The \textit{Inference} and \textit{Vector Store} components are responsible for managing communication with the LLM and the vector database, enabling efficient evaluation of RAG approaches.

For \textit{Inference}, we focus on local deployment, primarily leveraging \texttt{vLLM}~\cite{kwon_efficient_2023} as it represents one of the fastest LLM engines available. Communication with the LLM is handled through the OpenAI Python SDK~\cite{openai_openai-python_2025}, allowing seamless integration of HuggingFace, OpenAI, and Gemini. Other integrations are planned.

For the \textit{Vector Store}, we utilize a serverless version of Chroma~\cite{chroma_team_chroma_2025}, which creates an in-memory \texttt{SQLite3} database stored locally. As an alternative, we also support Qdrant~\cite{qdrant_team_qdrant_2025}, which is capable of handling large-scale document collections. Both implementations share a unified interface, simplifying the integration of additional vector database backends in the future.

\subsection{Metrics} \label{subsec:metrics}

Evaluating different RAG methods is a core feature of our library, implemented through the \textit{Metrics} module. 
Table~\ref{tab:metrics_overview} gives an overview of the most popular metrics of the library. To provide a transparent and comprehensive assessment, we include metrics for all components of RAG across multiple datasets, grouped into three categories: generator metrics, retrieval metrics, and LLM-based metrics. To align with existing research, we primarily use the \texttt{evaluate}~\cite{huggingface_evaluate_2025} library from HuggingFace for most generator metrics. For retrieval metrics, we used the \texttt{ir\_measures}~\cite{macavaney_streamlining_2022} library, which is specialized in information retrieval evaluation. 
LLM-based metrics are fully implemented and easily customizable with new prompts and examples, enabling flexible use across datasets.
The \textit{Type Manifest} typing system is designed to integrate seamlessly with all metrics, enabling automatic computation once a RAG method is selected.

\begin{table}[h!]
\centering
\small
\begin{tabular}{p{0.4\columnwidth} | p{0.55\columnwidth}}
\hline
\textbf{Metric } & \textbf{Description} \\
\hline 
\multicolumn{2}{l}{\textit{Generator Metrics}} \\
\hline 
\textbf{Exact Match}  & Checks if the generated text exactly matches the reference. \\
\parbox[t]{0.4\columnwidth}{\textbf{Number Match}\\\textcolor{darkblue}{(Anonym)}} & Checks if generated text is numerical aprox. equal.  \\[3ex]
\parbox[t]{0.4\columnwidth}{\textbf{BLEU}\\\cite{papineni_bleu_2002}} & Measures n-gram overlap between generated and reference text. \\[5ex]
\parbox[t]{0.4\columnwidth}{\textbf{ROUGE}\\\cite{lin_rouge_2004}} & Measures overlap of n-grams and sequences. \\[4ex]
\parbox[t]{0.4\columnwidth}{\textbf{GLEU}\\\cite{mutton_gleu_2007}} & Modified BLEU emphasizing recall and precision balance. \\[4ex]
\parbox[t]{0.4\columnwidth}{\textbf{Precision}\\\cite{sokolova_systematic_2009}} & Fraction of generated tokens that are correct. \\[5ex]
\parbox[t]{0.4\columnwidth}{\textbf{Recall}\\\cite{sokolova_systematic_2009}} & Fraction of reference tokens correctly generated. \\[5ex]
\parbox[t]{0.4\columnwidth}{\textbf{F1}\\\cite{sokolova_systematic_2009}} & Harmonic mean of precision and recall at token level. \\[5ex]

\hline
\multicolumn{2}{l}{\textit{Retrieval Metrics}} \\
\hline 
\parbox[t]{0.4\columnwidth}{\textbf{Mean Reciprocal Rank (MRR)}\\\cite{voorhees_using_1993}} & Mean reciprocal rank of the first relevant document retrieved. \\[5ex]
\parbox[t]{0.4\columnwidth}{\textbf{Mean Average Precision (MAP)}\\\cite{voorhees_using_1993}} & Mean of average precision scores across queries. \\[4ex]
\parbox[t]{0.4\columnwidth}{\textbf{nDCG}\\\cite{jarvelin_cumulated_2002}} & Normalized discounted cumulative gain, emphasizing top-ranked relevance. \\[6ex]
\parbox[t]{0.4\columnwidth}{\textbf{Recall@k}\\\cite{robertson_probabilistic_2009}} & Fraction of relevant documents retrieved in top-k. \\[6ex]
\parbox[t]{0.4\columnwidth}{\textbf{HitRate@k}\\\cite{li_multi-interest_2019}} & Measures if at least one relevant document appears in the top-k results. \\[2ex]
\hline
\multicolumn{2}{l}{\textit{LLM-Based Metrics}} \\
\hline 
\parbox[t]{0.4\columnwidth}{\textbf{Answer Relevance}} & Assesses if the answer addresses the query. \\[3ex]
\parbox[t]{0.4\columnwidth}{\textbf{Answer Faithfulness}} & Measures factual consistency with the context. \\[3ex]
\parbox[t]{0.4\columnwidth}{\textbf{Non-Answer Critic}} & Detects answers that fail to provide meaningful responses. \\[3ex]
\parbox[t]{0.4\columnwidth}{\textbf{Answer Similarity}} & Compares generated answer to reference or alternative outputs. \\[3ex]
\parbox[t]{0.4\columnwidth}{\textbf{Context Recall}} & Measures how well the retrieved context supports the answer. \\[3ex]
\parbox[t]{0.4\columnwidth}{\textbf{Context Precision}} & Fraction of retrieved context that is relevant to answer. \\[3ex]
\hline
\end{tabular}
\caption{Overview of the most used RAG evaluation metrics that we implemented in \textit{EncouRAGe}.}
\label{tab:metrics_overview}
\end{table}

\begin{table*}[!th]
\centering
\resizebox{\textwidth}{!}{
\begin{tabular}{l|l|cc|cc}
\toprule
\textbf{Subset} & 
\textbf{Domain} & 
\textbf{\#QA Pairs} &
\textbf{Avg. Tokens (Question)} &
\textbf{\#Documents} &
\textbf{Avg. Tokens (Docs)} \\
\midrule
BioSQA     & Biomedical         & 4,012 & 13.5 & 34,634 & 3,222.8 \\
FinQA      & Finance            & 6,237 & 45.1 & 2,110 & 1,080.4 \\
FeTaQA     & Table Questions    & 7,324 & 20.4 & 6,929 & 520.0  \\
HotPotQA   & General Questions  & 7,395 & 21.4 & 7,395 & 1,421.2 \\
\midrule
\textbf{Total} & \textbf{Multiple} & \textbf{24,968} & \textbf{25.7} & \textbf{51,068} & \textbf{2,506.7} \\
\bottomrule
\end{tabular}
}
\caption{
Comparison of QA pairs, documents, and average token lengths across subsets. HotPotQA~\cite{yang_hotpotqa_2018} \& FeTaQA~\cite{nan_fetaqa_2022} are based on Wikipedia articels, while FinQA~\cite{chen_finqa_2021} \& BioSQA~\cite{krithara_road_2023} uses PDF documents. Avg. token count based on Gemma-3 tokenizer.
}
\label{tab:dataset_stats}
\end{table*}

\section{Experiments} \label{sec:experiments}

To demonstrate the effectiveness of our library, we conduct a comprehensive analysis of four popular datasets using each RAG method implemented in EncouRAGe. The datasets used are summarized in Table~\ref{tab:dataset_stats}. Each QA pair has at least one associated golden document, enabling comparisons of RAG methods across pretrained, oracle, and various RAG settings. We evaluate a total of nine RAG methods across all datasets, as presented in Table~\ref{tab:main_results_full}, and discuss the results in Subsec.~\ref{subsec:main_results}.

Additionally, we conduct ablation experiments on the reranker models in Subsec.~\ref{subsec:reranker}.  
We define the reranker ratio as the proportion of initially retrieved documents that are reranked in the second phase, after which a fixed set of 10 papers is provided to the generator model.  
This should evaluate how different reranker ratios affect BaseRAG’s performance and whether combining it with rerankers yields further improvements.

\subsection{Experimental Setup} \label{subsec:experimental_setup}

For the evaluation of the dataset, each subset was processed and evaluated independently. First, all contexts were uniquely stored in a vector database using embeddings created with the multilingual e5-large instruct model~\citep{wang_multilingual_2024}. That was done for all RAG methods except for the Summarization, where the summarized context was embedded. A retrieval query was used to retrieve the context from the instruction model. The Top-10 documents were passed to each method of the \textit{RAG Factory} and passed to the \textit{Inference} component in the main evaluation. As generators, we employed Gemma3 27b~\cite{kamath_gemma_2025}, a SOTA decoder-only transformer. All experiments were conducted on two NVIDIA H100s.

\subsection{Evaluation Metrics} 
We leveraged EncouRAGe's flexibility to evaluate each dataset using task-specific metrics tailored to the required format. For datasets where only a single document contains the gold answer, we employed \textit{MRR@10} to measure the position of the retrieved document. For tasks requiring multiple documents to derive the correct answer, we used \textit{MAP}, in line with prior studies. Additionally, we report \textit{Recall@10} to assess overall retrieval performance, complementing \textit{MRR} by considering the rank of retrieved documents. This is particularly informative in the reranker setup.

For the generation, we also used multiple metrics, depending on the task. For short answers, we used \textit{F1-Score}, for numerical answers, we used \textit{NM}~\textcolor{darkblue}{(Anonym)}.

\begin{table*}[!th]
\centering
\scriptsize
\resizebox{\textwidth}{!}{
\begin{tabular}{>{\raggedright}p{1.5cm}l|ccc|ccc|ccc|ccc}
\toprule
\multirow{2}{*}{\textbf{Model}} & \multirow{2}{*}{\textbf{RAG Method}}
  & \multicolumn{3}{c}{\textbf{HotPotQA}}
  & \multicolumn{3}{c}{\textbf{FeTaQA}}
  & \multicolumn{3}{c}{\textbf{FinQA}}
  & \multicolumn{3}{c}{\textbf{BioSQA}} \\
\cmidrule(lr){3-5} \cmidrule(lr){6-8} \cmidrule(lr){9-11} \cmidrule(lr){12-14}
& & F1 & MRR & R@10 & F1 & MRR& R@10 & NM & MRR & R@10 & F1 & MAP & R@10 \\
\midrule
\multirow{8}{*}{\centering\makecell{Gemma 3 27B\\+ M-E5-Large\\Instruct}}
 & + Pretrained-Only & 36.7 & -- & -- & 29.3 & -- & -- & 9.6 & -- & -- & 39.3 & -- & -- \\
 & + Oracle Context  & 43.4 & 100 & 100 & 49.4 & 100 & 100 & 72.9 & 100 & 100 & 54.5 & 100 & 100 \\
 \cmidrule(lr){2-14}
 & + Base‐RAG        & 37.1 & 68.8 & 82.5 & 49.8 & 87.5 & \underline{\textbf{92.7}} & 47.8 & 45.6 & 71.9 & 47.2 & 42.1 & 50.1 \\
 & + BM25            & 40.5 & 29.1 & 98.4 & 39.0 & 18.6 & 68.5 & 50.8 & 45.5 & 77.2 & 44.3 & 26.7 & 44.3 \\
 & + Hybrid BM25     & \underline{\textbf{40.8}} & \underline{70.4} & \underline{\textbf{96.9}} & \underline{\textbf{50.1}} & \underline{\textbf{87.6}} & 92.4 & \underline{\textbf{51.8}} & \underline{46.7} & \underline{\textbf{82.1}} & \underline{\textbf{49.9}} & \underline{43.6} & \underline{\textbf{55.7}} \\
 & + Reranker        & 36.1  & 63.4 & 78.4 & 49.6 & 86.9 & 92.2 & 50.2 & 45.5 & 71.7 & 44.5 & 42.1 & 50.1 \\
 \cmidrule(lr){2-14}
 & + HyDE            & 30.2 & 40.6 & 61.0  & 48.4 & 82.8 & 89.5  & 41.7 & 37.3 & 61.4  & \underline{49.2} & \underline{\textbf{45.9}} & \underline{54.7} \\
 & + Summarization   & 31.7 & \underline{\textbf{71.0}} & \underline{83.3}  & 40.0 & \underline{83.9} & \underline{90.5}  & 45.9 & \underline{\textbf{48.5}} & \underline{74.6} & 45.0 & 43.2 & 51.0 \\
 & + SumContext      & \underline{37.7} & \underline{\textbf{70.9}} & \underline{83.3}  & \underline{48.6} & \underline{84.0} & \underline{90.5}  & \underline{48.6} & 48.1 & \underline{74.6} & 47.8 & 43.0 & 50.4  \\
\bottomrule
\end{tabular}
}
\caption{%
Overall Performance (F1, MRR/MAP, R@10) for Gemma3 with E5-Large Instruct on all four datasets.    
Cells in \textbf{Bold} indicate the highest value over all RAG methods,
and \underline{underlined} indicate the best value across RAG method categories.
}
\label{tab:main_results_full}
\end{table*}

\subsection{Datasets} \label{subsec:datasets}

For evaluation, we selected four datasets from different domains to demonstrate the flexibility of EncouRAGe in handling diverse datasets as presented in Table~\ref{tab:dataset_stats}.

\paragraph{HotPotQA.} 
The dataset~\citep{yang_hotpotqa_2018} contains 7,395 general knowledge question–answer pairs sourced from Wikipedia, emphasizing multi-hop reasoning. Each question is linked to a single gold document, resulting in 7,395 papers in total, with an average document length of 1,421.2 tokens. Questions average 21.4 tokens, reflecting concise query formulation. HotPotQA challenges models to combine information across multiple articles to answer a single question accurately. For evaluation, we employ \textit{MRR@10} for retrieval and \textit{F1-Score} for answer generation, consistent with multi-hop 
reasoning tasks.

\paragraph{FetaQA.}
The dataset~\citep{nan_fetaqa_2022} includes over 7,300 QA pairs that require complex reasoning over tabular data, extracted from Wikipedia articles. The dataset emphasizes semantic understanding and reasoning across multiple table cells. Each question is accompanied by a textual explanation and a gold-standard answer derived from structured tables. We use the training split and all table documents related to these QA pairs, resulting in a total of 6,929 documents to process.  Evaluation employs \textit{MRR@10}, \textit{Recall@10}, and \textit{F1-Score} to measure both retrieval effectiveness and answer quality.

\paragraph{FinQA.}
The dataset~\citep{chen_finqa_2021} comprises approximately 6,200 finance-related question–answer pairs derived from company reports and other financial documents. Each question is associated with a single financial document containing a combination of text and tables. On average, there are roughly three QA pairs per document, resulting in a total of 2,110 documents for retrieval. 
We used a modified version of the training split, reformulating questions to make them more suitable for RAG. For reasons of anonymity, the original paper is not cited in this submission. For evaluation, we employ \textit{MRR@10}, \textit{Recall@10}, and \textit{Number Match (NM)}, a metric specifically designed to compare numerical values, like Exact Match, up to the second decimal place.

\paragraph{BioASQ.}
The dataset~\citep{krithara_road_2023} contains approximately 4,000 biomedical question–answer pairs derived from scientific literature in the life sciences. It is designed to evaluate factual biomedical question answering in combination with semantic retrieval. The dataset is updated annually, and the version used in our evaluation is BioASQ11. Each QA pair includes multiple reference documents, over 34k in total, making it particularly challenging for the retriever to locate the relevant information.  
Questions are grouped into four types: list, yes/no, factoid, and summary, each requiring different response formats. For evaluation, we use \textit{MAP} and \textit{Recall@10} for retrieval, and \textit{F1-Score} for generation.




\section{Main Results}\label{subsec:main_results}

Table~\ref{tab:main_results_full} presents the quantitative results across the each of the four QA datasets using each RAG method from Sec.~\ref{subsec:rag_factory}. 

\paragraph{Without RAG Runs.}  
The \textit{Pretrained-Only} results for HotPotQA, FeTaQA, and BioSQA are comparatively high, indicating that these datasets are likely part of the LLM’s pretraining corpus (e.g., Wikipedia-based sources). In contrast, FinQA shows a substantially lower score, suggesting that its domain-specific financial questions are less represented in the model’s training data. However, when provided with \textit{Oracle Context}, all datasets achieve higher performance than in the \textit{Pretrained-Only}, confirming that the model can effectively reason when supplied with relevant context. Notably, the \textit{Oracle Context} results for HotPotQA~\cite{yang_hotpotqa_2018} and FeTaQA~\cite{nan_fetaqa_2022} are slightly below the original papers, as our setup relies solely on a zero-shot setting.

\paragraph{Basic RAG.}  
When incorporating retrieval \textit{Base-RAG} performs close to the \textit{Oracle Context} for HotPotQA, FeTaQA, and BioSQA, while a larger performance gap appears for FinQA. Interestingly, the traditional \textit{BM25} retriever performs comparably to the dense retriever in terms of F1, but exhibits substantially lower MRR on HotPotQA and BioSQA. Combining both retrieval methods (\textit{Hybrid BM25}) yields the most consistent improvements, achieving the highest F1 and Recall@10 across all datasets. The addition of a reranker does not lead to measurable gains, as the MRR remains similar to the base configuration when reordering the top-10 retrieved documents.
We discuss the effect of the Reranker further in Subsec.~\ref{subsec:reranker}.

\paragraph{Advanced RAG.}  
Among the advanced retrieval enhancements, \textit{HyDE} offers modest improvements only for BioSQA, which are insufficient to justify its additional computational cost. \textit{Summarization}-based retrieval increases MRR and Recall@10 for all datasets except FeTaQA, but leads to lower F1 scores, likely due to information loss during summarization. The combined approach, \textit{SumContext}, which integrates both summarized and original documents, achieves a balanced trade-off, yielding results similar to \textit{Base-RAG} with slightly better retrieval performance. However, given its additional retrieval and generation overhead, its practical benefit remains limited.

Overall, the results demonstrate that hybrid retrieval remains the most effective and efficient RAG method across diverse QA domains. In contrast, advanced generation-based retrieval strategies offer only marginal improvements at a higher computational cost.

\begin{figure*}[!t]
    \centering
    \includegraphics[width=1\textwidth]{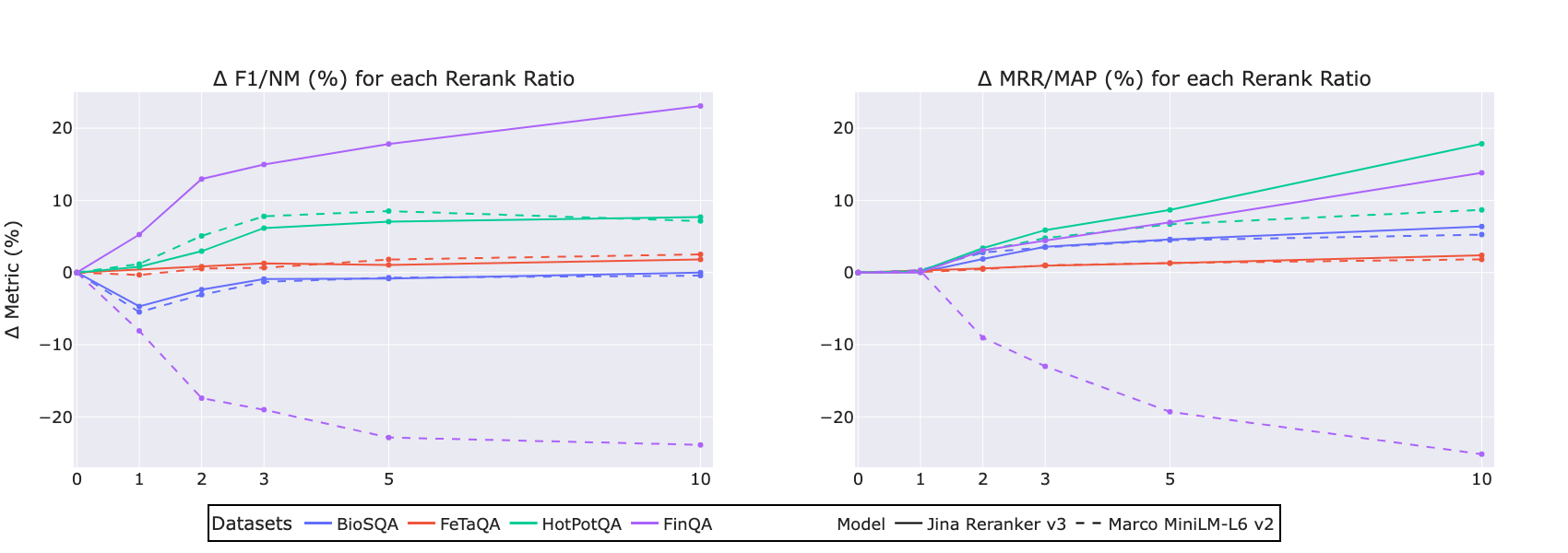}
    \caption[reranker]{Comparison of \textbf{percentage changes for generator (F1/NM) and retrieval metrics (MRR/MAP)} for a 2k-sample subset from each dataset (HotPotQA, FeTAQA, FinQA, and BioSQA). Reranking was performed using Jina v3 and Marco MiniLM-L6 v2. The x-axis represents the reranker ratio, and the y-axis shows the percentage change relative to the base RAG method.}
    \label{fig:reranker}
\end{figure*}

\subsection{Reranker Analysis}\label{subsec:reranker}
We further analyzed reranker performance using \textit{EncouRAGe}. A widely discussed claim is that rerankers consistently improve retrieval and generation quality independent of the domain and data format of the documents. Most approaches suggest retrieving a broader set of documents and reranking them before passing the top results to the generator~\cite{clavie_rerankers_2024, bhat_ur2n_2025}.

To investigate this effect, we sampled 2k QA pairs from each of the four datasets and applied two reranker models: Jina V3~\cite{sturua_jina-embeddings-v3_2024} and Marco MiniLM-L6 v2~\cite{reimers_making_2020}, with five reranker ratios, and compared them against the Base RAG method. For each run, 10 documents were provided to the generator model (Gemma3 27B). A reranker ratio of 1 means only these 10 documents were reranked, while a ratio of 10 indicates that 100 papers were initially retrieved, reranked, and the top 10 were passed to the generator. Evaluation metrics for both generator and retriever followed the same setup as above.

Figure~\ref{fig:reranker} shows the percentage change from the baseline (Base RAG) for generator metrics (F1/NM) and retriever metrics (MRR/MAP) across reranker ratios 1–10. 
We observe consistent performance gains in the retriever metrics (except for FinQA + Marco), but only meaningful gains in HotPotQA and FinQA. For the generator metrics, even as retrieval performance improved, saturation was observed at a ratio of 3 (except for FinQA + Jina).
For FinQA, we hypothesize that the reranker, unlike Jina V3 trained on FinQA samples, was not trained on mixed text–table data, leading to noisy, random rankings.

For FeTaQA, which uses table-based contexts, neither reranker provided notable improvements. This suggests that rerankers offer benefits only when explicitly trained for the data format. Finally, while cross-encoders outperform bi-encoders, reranking increased execution time by 2–4× over the base RAG method, which may pose challenges for production systems that require low-latency responses. Therefore, we recommend using a reranker model only when it is individually trained on the dataset and when high throughput is not the key metric.

\vspace{-0.3cm}
\section{Conclusion}
In this paper, we introduce \textbf{EncouRAGe}, a Python framework for evaluating RAG methods locally, reliably, and efficiently. EncouRAGe enables researchers to systematically investigate retrieval-augmented generation techniques on their own datasets, providing insights into which configurations yield the best results for specific domains. The framework includes over 10 RAG methods, more than 20 evaluation metrics, and supports models from Hugging Face, OpenAI, and Gemini. Furthermore, users can flexibly choose between Chroma and Qdrant as the vector store backend.

To demonstrate the capabilities of EncouRAGe, we conducted experiments on four widely used datasets, revealing that the optimal RAG method varies across domains. Overall, our results indicate that the \textit{Hybrid BM25} approach delivers the best performance among the tested configurations. Additionally, we performed an ablation study on reranker models to examine the influence of the reranker ratio. Our findings show that increasing this ratio can boost performance by up to 10\%, even if at the cost of higher latency, a trade-off that should be considered when deploying rerankers in real-world systems.

With EncouRAGe, we aim to contribute to the RAG research community by providing a flexible, extensible evaluation framework that enables deeper analysis and supports the development of practical, high-performing systems.

\clearpage

\section{Bibliographical References}\label{sec:reference}

\bibliographystyle{lrec2026-natbib}
\bibliography{Encourage}

\end{document}